# HHTRACK: HYPERSPECTRAL OBJECT TRACKING BASED ON HYBRID ATTENTION

*Yuedong Tan[1], Wenfang Sun[1,*], Jieran Yuan[2], Wenwang Du[2], Zhe Wang[2], Nan Mao[2], Beibei Song[2,*]*

[1]School of Aerospace Science and Technology, Xidian University, Xi'an 710126, China
[2]School of Information Engineering, Chang'an University, Xi'an 710064, China

**ABSTRACT**

Hyperspectral imagery provides abundant spectral information beyond the visible RGB bands, offering rich discriminative details about objects in a scene. Leveraging such data has the potential to enhance visual tracking performance. In this paper, we propose a hyperspectral object tracker based on hybrid attention (HHTrack). The core of HHTrack is a hyperspectral hybrid attention (HHA) module that unifies feature extraction and fusion within one component through token interactions. A hyperspectral bands fusion (HBF) module is also introduced to selectively aggregate spatial and spectral signatures from the full hyperspectral input. Extensive experiments demonstrate the state-of-the-art performance of HHTrack on benchmark Near Infrared (NIR), Red Near Infrared (Red-NIR), and Visible (VIS) hyperspectral tracking datasets. Our work provides new insights into harnessing the strengths of transformers and hyperspectral fusion to advance robust object tracking.

## 1. INTRODUCTION

Visual object tracking [1] refers to continuously tracking a target after determining the object to be tracked. Object tracking has been widely used in sports broadcasting, security surveillance, and fields involving unmanned aerial vehicles, driverless cars, robots, etc. However, RGB-based tracking can only provide relatively limited information and struggles with complex backgrounds, occlusions, etc. Hyperspectral images (HSI) can provide much richer spectral details compared to RGB images. Therefore, HSI-based tracking has great potential to improve tracking performance. Hyperspectral trackers can be roughly categorized into two types: correlation filter-based trackers and Siamese network-based trackers.

DeepHKCF [2] uses convolutional neural networks (CNN) combined with correlation filters for tracking. MHT [3] integrates local spectral-spatial histogram of multidimensional gradients (SSHMG) features and BACF [14] for tracking. TSCFW [4] proposes a reformulated stacked histogram of oriented gradient (SHOG) feature and uses a template update strategy with correlation filters for tracking. Correlation filter-based methods suffer from inductive bias, making them struggle with complex scenes.

Siamese network-based approaches can effectively extract features from both the template and search regions. SiamHYPER [5] utilizes the 1st, 9th and 16th channels of the hyperspectral image as well as the full hyperspectral image as input to a dual Siamese network. BAE-Net [6] proposes a channel attention module to group channels into weak trackers. BS-SiamRPN [7] introduces a band selection strategy with SiamRPN [8] for tracking. SiamBAG [9] also employs channel grouping by dividing channels into multiple RGB inputs fed into the Siamese network. By incorporating transformers, TMTNet [10] leverages self-attention and cross-attention to sufficiently extract and fuse features from the three Siamese branches. Similarly, SSTFT [11] uses two CNNs for feature extraction followed by an attention mechanism for feature fusion. CBFF-Net [12] divides hyperspectral image channels into five groups. It fuses the group features using bidirectional fusion and cross-band attention modules, and then inputs them to a transformer fusion module.

In this paper, we propose a novel transformer-based hyperspectral tracker called HHTrack. Compared to previous Siamese network-based trackers, our HHA module interacts with tokens at every stage to jointly extract and fuse features, instead of fusing only at the last stage. Similar to SiamHYPER, we utilize both 3-channel and full spectral images to alleviate the issue of limited training data. Furthermore, to sufficiently extract and integrate target information, we design an HBF module to better fuse spectral details from the full HSI input.

## 2. METHODOLOGY

Mixformer [13] employs mixed attention to fuse features from the template and search regions. To better utilize hyperspectral information for enhanced tracking, HHTrack incorporates spectral details into the attention mechanism. By interacting false-color images sampled at equal intervals in channel dimension and the entire spectral image, more robust features can be extracted. In this section, we first present an overview of the HHTrack architecture, and then detail the HHA module and HBF module respectively.

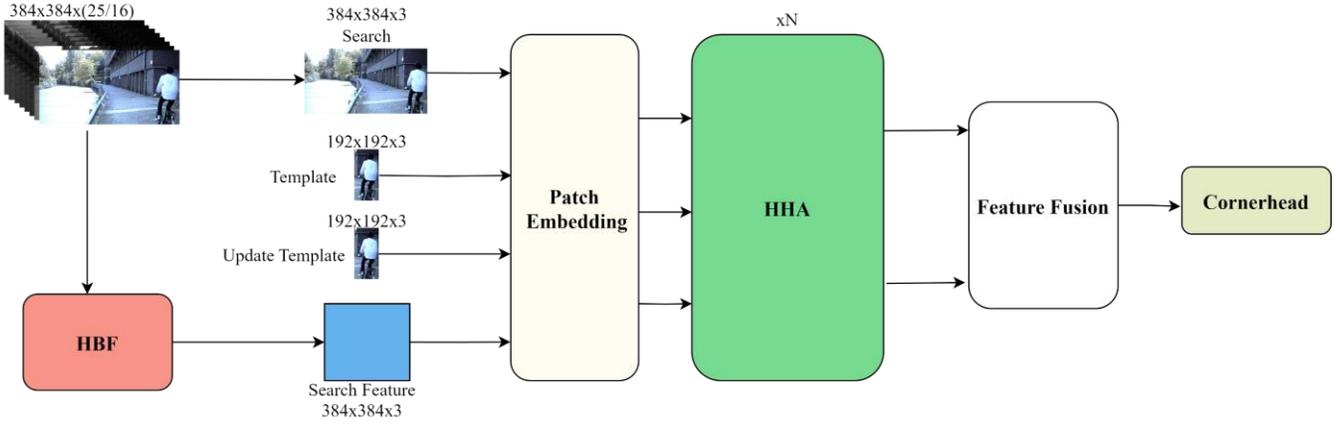

**Fig.1.** Our HHTrack framework. The search representation and templates are projected to transformer embeddings and interact via multi-branch attention in the HHA.

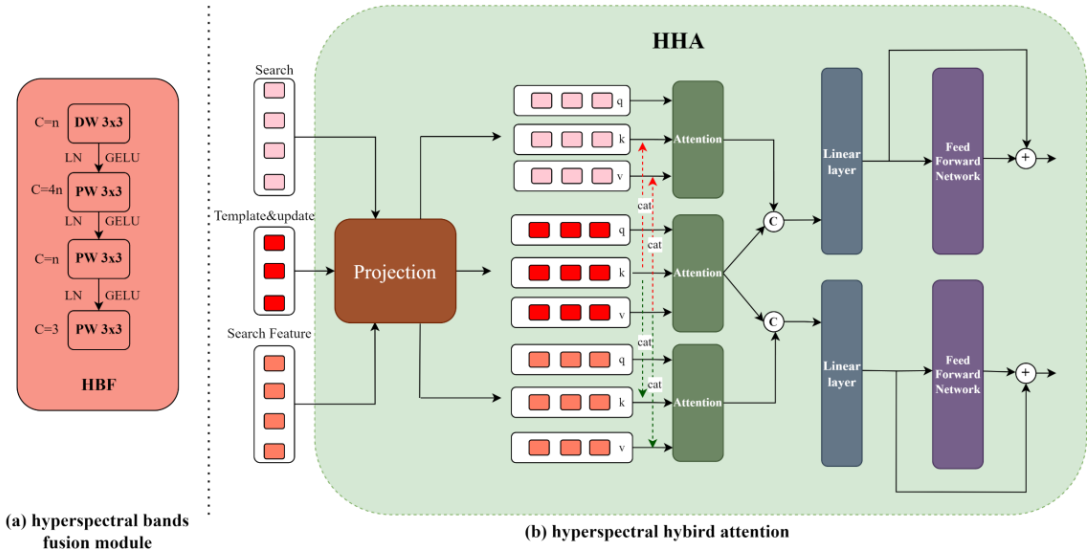

**Fig.2.** (a) HBF module is designed to effectively process the full hyperspectral search region input.
(b) HHA is the core design of HHTrack. HHA employs a multi-branch self-attention mechanism to enable interactions between the template representation and search region feature.

### 2.1. Overall architecture

HHTrack adopts a one-stream encoder-only transformer architecture as shown in Fig. 1. It consists of the HBF, patch embedding, HHA, and corner head. First, the search region is divided into two parts, one for the false-color input by sample 3 bands with the same interval and the full hyperspectral image fed into the HBF. The HBF performs joint feature extraction and fusion, aggregating information across channels while reducing redundancy. For the template and updated template, we adopt the same sampling strategy as the Search to sample three channels as the input. After shared patch embedding, they are input into the HHA module. Attention, we concatenate tokens from the template and the updated template. The HHA integrates spectral features from the search region and false-color search with the template features. A simple maximal fusion will be applied to the extracted features, where the maximal fusion takes the element-wise maximum between two features of the same size to generate the new feature. Finally, similar to [19], we utilize corner heads to predict the position of the target.

## 2.2. HBF Module

As shown in Fig.2 (a), the HBF comprises one Depthwise (DW) convolution followed by three Pointwise (PW) convolutions for effective spatial-spectral fusion. First, a 3x3 DW convolution extracts low-dimensional features from the hyperspectral search region. This maintains spatial details while reducing the spectral dimensionality. Next, three successive PW convolutions aggregate information across the spectral channels in a pixel-wise manner. The forward pass of DW convolution and PW convolution can be formulated as follows:

$$F' = GELU(LN(dwconv(F))) \quad (1)$$

$$F' = GELU(LN(pwconv(F))) \quad (2)$$

where $F$ denotes the original hyperspectral image or the high-dimensional feature, and $F'$ represents the result of output layers.

The number of output channels is also reduced to 3, forming a compact and discriminative spectral-spatial representation. The DW-PW convolution design allows dense cross-channel interactions to select useful spectral bands and fuse them in spectral dimension. By adaptively emphasizing informative spectral signatures and spatial patterns, the HBF module produces a robust fused representation from the full hyperspectral input. This complements the false-color images to provide comprehensive target details for precise tracking.

## 2.3. Hyperspectral Hybrid Attention

The proposed Hyperspectral Hybrid Attention (HHA) module concurrently extracts and fuses features for the target representation. After passing through the shared patch embedding, the initial template tokens and updated template tokens (template tokens), search tokens, and search feature tokens are linearly projected to obtain their respective queries, keys, and values. To enable feature interactions, we concatenate the template keys and values to those of the search and search features respectively. Let $q_s$, $k_s$, $v_s$ denote the query, key, and value embeddings obtained from the linear projection of the search tokens. Similarly, $q_f$, $k_f$, $v_f$ and $q_t$, $k_t$, $v_t$ represent the projected embeddings of the search feature tokens and template tokens respectively. We concatenate $k_s$ with $k_t$ and $v_s$ with $v_t$ to obtain $k_a$ and $v_a$. Likewise, $kf$ is concatenated with $k_t$ while $v_f$ with $v_t$ to produce $k_b$ and $v_b$. As elaborated, the three attention branches in the HHA module are defined as follows:

$$Attention_t = Soft\max\left(\frac{q_t k_t^T}{\sqrt{d}}\right)v_t \quad (3)$$

$$Attention_s = Soft\max\left(\frac{q_s k_a^T}{\sqrt{d}}\right)v_a \quad (4)$$

$$Attention_f = Soft\max\left(\frac{q_f k_b^T}{\sqrt{d}}\right)v_b \quad (5)$$

where $d$ represents the dimension of the key, value, and query. $Attention_t$, $Attention_s$ and $Attention_f$ are the attention maps of the template, search, and search features respectively. After applying attention in three parallel branches, the template tokens are added to the search and search feature token followed by a feedforward network that consists of two linear layers. In essence, the HHA interacts with the template, search, and search feature representations via multi-branch attention. This allows for enriching the template embedding with complementary information extracted from the search region in an iterative manner. The concurrent extraction and fusion of target and search features is more effective than separate processing. The detailed structure of HHA is depicted in Fig.2(b).

## 3. EXPERIENCES

### 3.1. Implementation Details

To evaluate HHTrack, we utilize the training and validation sets provided in this competition [15], comprising NIR, NIR-Red, and VIS data. The training set contains 40 NIR videos, 14 NIR-Red videos, and 55 VIS videos. The validation set has 30 NIR, 11 NIR-Red, and 46 VIS videos. We initialize HHTrack with pre-trained weights from [13]. For training, the ADAMW optimizer is used with CIoU loss. The process involves two stages. First, the backbone and head are frozen and only the HBF is trained for 20 epochs at a learning rate of 0.0004. Second, the backbone and HBF are jointly optimized for 5 epochs at a reduced learning rate of 0.0001. HHTrack is implemented on an Intel Xeon CPU with an Nvidia RTX3090 GPU. During testing, the template and an update template sampled every 25 frames along with the search region are fed as inputs to the network. The update template facilitates adapting to appearance changes over time.

### 3.2. Qualitative Comparison

To validate our approach, we compare HHTrack with recent RGB trackers including TransT [16], SiamGAT [17], SiamCAR [18], STARK [19], and SiamBAN [20]. The RGB trackers are evaluated on false-color videos derived from the NIR, Red-NIR, and VIS hyperspectral datasets respectively, while HHTrack is tested on the original hyperspectral videos. Attention, the number of input channels to the HBF module differs for the three hyperspectral data types. We utilize precision plots, success plots, and area under the curve (AUC) of one-pass evaluation (OPE) to assess all trackers. As shown in Fig. 3, HHTrack achieves a DP score of 0.817 and AUC score of 0.817, exceeding the baseline RGB trackers by 5.4% and 4.0% respectively.

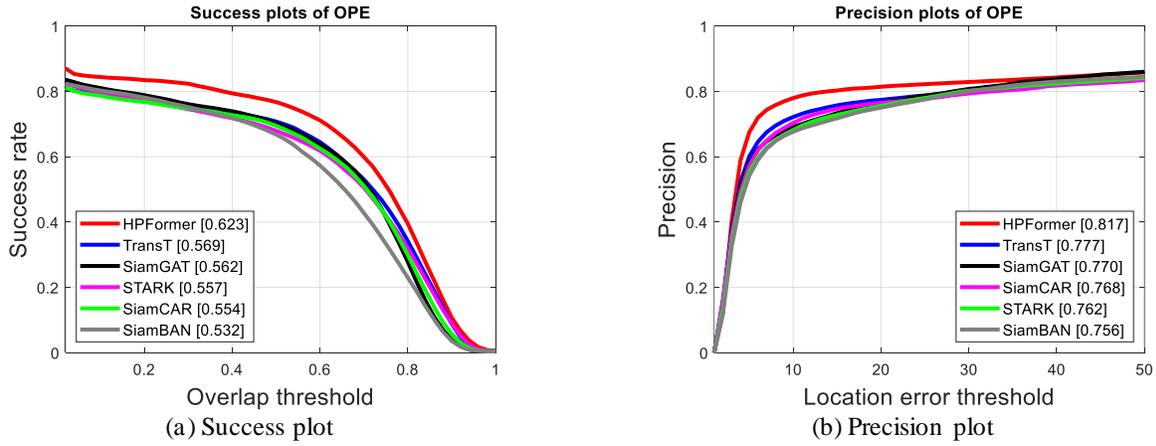

(a) Success plot  (b) Precision plot

**Fig.3.** The success plot and precision plots of all trackers

**Table 1.** The AUC/DP scores of the HHTrack and RGB-based Trackers on false-color or hyperspectral image(validation dataset)

| Trackers | HHTrack | TransT | SiamGAT | STARK | SiamCAR | SiamBAN |
|---|---|---|---|---|---|---|
| **AUC/DP** | 0.623/0.817 | 0.569/0.777 | 0.562/0.770 | 0.557/0.768 | 0.554/0.762 | 0.532/0.756 |

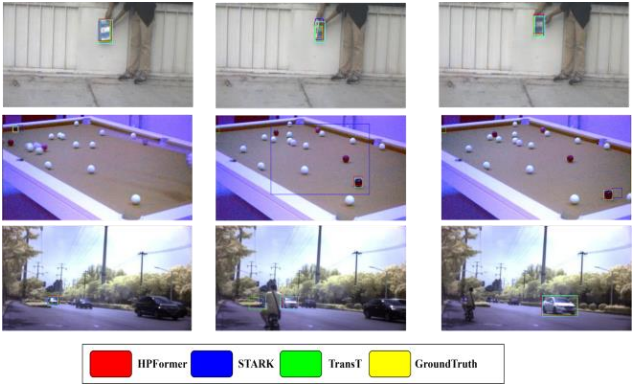

**Fig.4.** Visualization results of different methods

## 4. CONCLUSION

In this work, we have introduced a novel transformer-based architecture for hyperspectral visual tracking called HHTrack. By incorporating spectral information into the attention mechanism via the proposed Hyperspectral Hybrid Attention module, more robust target representations can be extracted through joint feature extraction and fusion. To fully utilize the hyperspectral input, a HBF module is designed to selectively aggregate spatial details and spectral signatures. Extensive experiments on benchmark hyperspectral datasets demonstrate that HHTrack significantly outperforms state-of-the-art RGB trackers. By effectively harnessing spectral cues, our approach produces more discriminative embeddings to handle common tracking challenges such as occlusion and deformation. This validates the benefits of leveraging hyperspectral imagery to enhance tracking performance. Our work provides valuable insights into effectively harnessing the strengths of transformers and hyperspectral data for robust visual tracking. Our future research will focus on developing a single model to track different hyperspectral modalities (NIR, Red-NIR, VIS) using a shared representation.

## ACKNOWLEDGMENTS
This work was supported by the National Natural Science Foundation of China (grant number 62375211).